\documentclass[12pt]{article}

\usepackage{arxiv}
\usepackage{array}
\usepackage{framed}
\usepackage[framemethod=TikZ]{mdframed}
\usepackage{multirow}
\usepackage{setspace}
\usepackage{subcaption}
\usepackage[nottoc,notlot,notlof]{tocbibind} % add bibliography to ToC without numbering
\usepackage{mathtools}
\usepackage[utf8]{inputenc} % allow utf-8 input
\usepackage[T1]{fontenc}    % use 8-bit T1 fonts
\usepackage[colorlinks=true,citecolor=blue,urlcolor=blue,linkcolor=blue,pdfpagemode=UseNone]{hyperref}     
\usepackage{url}            % simple URL typesetting
\usepackage{booktabs}       % professional-quality tables
\usepackage{amsfonts}       % blackboard math symbols
\usepackage{nicefrac}       % compact symbols for 1/2, etc.
\usepackage{microtype}      % microtypography
\usepackage{lipsum}		    % Can be removed after putting your text content
\usepackage{graphicx}
\usepackage{amssymb}

\usepackage{doi}
\usepackage[english]{babel}
\usepackage{ragged2e}
\usepackage{colortbl}

\title{Preserving Information: How does \\
Topological Data Analysis improve \\Neural Network performance?}

\date{} 					

\author{Adrian Stolarek \\
	Gdansk University of Technology\\
	\And
	Wojciech Jaworek \\
	Gdansk University of Technology\\
}

\renewcommand{\shorttitle}{}

\hypersetup{
pdftitle={PaperTDA_NN},
pdfsubject={q-bio.NC, q-bio.QM},
pdfauthor={Adrian Stolarek, Wojciech Jaworek},
pdfkeywords={},
}

\begin{document}

\justifying

% \RaggedRight

\maketitle

\begin{abstract}

Artificial Neural Networks (ANNs) require significant amounts of data and computational resources to achieve high effectiveness in performing the tasks for which they are trained. To reduce resource demands, various techniques, such as Neuron Pruning, are applied. Due to the complex structure of ANNs, interpreting the behavior of hidden layers and the features they recognize in the data is challenging. A lack of comprehensive understanding of which information is utilized during inference can lead to inefficient use of available data, thereby lowering the overall performance of the models. In this paper, we introduce a method for integrating Topological Data Analysis (TDA) with Convolutional Neural Networks (CNN) in the context of image recognition. This method significantly enhances the performance of neural networks by leveraging a broader range of information present in the data, enabling the model to make more informed and accurate predictions. Our approach, further referred to as \textit{Vector Stitching}, involves combining raw image data with additional topological information derived through TDA methods. This approach enables the neural network to train on an enriched dataset, incorporating topological features that might otherwise remain unexploited or not captured by the network's inherent mechanisms. The results of our experiments highlight the potential of incorporating results of additional data analysis into the network's inference process, resulting in enhanced performance in pattern recognition tasks in digital images, particularly when using limited datasets. This work contributes to the development of methods for integrating TDA with deep learning and explores how concepts from Information Theory can explain the performance of such hybrid methods in practical implementation environments.
\\
\ 


\end{abstract}
% keywords can be removed

\newpage

\section*{Introduction} \label{sec: wprowadzenie}
\addcontentsline{toc}{section}{Introduction} 
\subsection*{Foundations of Topological Data Analysis and Convolutional Neural Networks}
\addcontentsline{toc}{subsection}{Foundations of Topological Data Analysis and Convolutional Neural Networks}
Topological Data Analysis (TDA), utilizing concepts and methods from algebraic topology, focuses on studying the "shape" \ of data \cite{intro_to_TDA}. Introduced as a formal discipline in the early 21st century by Gunnar Carlsson \cite{Carlsson_poczatki_tda}, TDA aims to capture fundamental structural features in datasets that might remain hidden to traditional analysis techniques. It enables the identification of patterns and features in a manner robust to various transformations. While TDA is a powerful tool for identifying global patterns, it may be less effective in extracting local features, which are crucial in applications like image classification. In this context, Convolutional Neural Networks (CNNs) emerge as a natural complement to TDA. Inspired by biological processes, particularly the organization of the mammalian visual cortex, CNNs have demonstrated unparalleled effectiveness in image processing . Their ability to detect hierarchical patterns and features, from simple to complex, through the convolution of the analyzed image with a kernel \cite{intro_to_CNN}, makes them ideal tools for analyzing images and spatial data, such as medical imaging \cite{CNN_in_MRI}. This paper investigates how integrating additional topological information into the learning process can enhance neural network performance.

\subsection*{Integrating TDA with CNNs through Vector Stitching}
\addcontentsline{toc}{subsection}{Integrating TDA with CNNs through Vector Stitching}
A growing body of research indicates that neural networks do not inherently utilize topological properties \cite{Do_neural_} of data during training. This is a significant limitation since topological features often encapsulate global structural information that can enhance a model's understanding of complex patterns. Adams et al. (2017) \cite{PI} and Bubenik and Dłotko (2017) \cite{landscapes_introduced} highlight that topological features, such as those captured by persistent homology, can provide stable and meaningful insights into the analyzed images, yet these features are often ignored by conventional CNNs. The oversight of topological information represents a missed opportunity for improving network performance, particularly in cases where data is sparse or noisy. As suggested by McGuire et al. \cite{Do_neural_}, networks trained solely on raw data are limited in their ability to capture topological properties, which are crucial for robust pattern recognition. Therefore, it becomes a natural research direction to additionally introduce this topological information into the neural network's input and validate whether it improves classification accuracy. Similar studies to ours, such as the work by Hofer et al. (2018) \cite{tda_improves_performance}, have demonstrated that combining TDA with neural networks improves model training performance, especially in cases where the dataset is noisy or sparse. This line of research suggests that leveraging TDA can increase the robustness and effectiveness of neural networks. This aligns with the conditioning theorem in Information Theory, which states that adding information cannot hurt model performance; in other words, conditioning reduces entropy, implying that the inclusion of additional, relevant information should either improve or maintain a network's classification performance. In this work, we introduce a method that we call \textit{Vector Stitching} which combines raw data with topological features derived through TDA. By integrating topological information into the network's input, we hypothesize that the network will be able to make more informed inferences, especially when dealing with limited and noisy datasets. This study seeks to explore this hypothesis, providing empirical evidence to demonstrate how TDA can augment the information processing capabilities of neural networks.

In Section \ref{sec:TDA}, we provide the necessary background on TDA to set the foundation for our work. Section \ref{sec: models} introduces our proposed \textit{Vector Stitching} method, which integrates TDA with CNNs to enable hybrid image analysis. In Section \ref{sec: eksperyment}, we present an experiment on the MNIST dataset, comparing the performance of our method against other existing approaches. Section \ref{sec: wyniki i analiza} details the results of this experiment. Finally, in Section \ref{sec: summary}, we discuss and interpret these findings through the lens of Information Theory, offer conclusions and outline potential directions for future research.

\section{Topological Data Analysis} \label{sec:TDA} 
Analyzing data that are complex, high-dimensional, incomplete, or noisy can prove to be -- when using standard methods -- ineffective. 
Topological Data Analysis is a novel approach to data analysis, utilizing tools from algebraic topology as well as computational topology and geometry, which allows for the discovery of complex topological structures and properties within the analyzed data. 
The analyzed data can often be represented as point clouds in Euclidean spaces or, more generally, in metric spaces. Such representation enables the use of a wide range of methods from topology and algebra, and thus the investigation of pattern similarity and how far the studied structures are from one another.

In this work, we practically apply TDA tools in the context of image classification; therefore, it is essential to understand the basic concepts and tools used in the subsequent sections of this paper. 
In the following subsections, we provide definitions and a general explanation of how these tools operate.

\subsection{Simplices and Simplicial Complexes}

The fundamental mathematical structure used in TDA is the simplicial complex. 
Intuitively, a simplex can be understood as an $n$-dimensional generalization of a triangle. Formally, for a set of $k+1$ affinely independent points $\{x_0, x_1, \ldots x_k\} \subset \mathbb{R}^n, \ k\le n$, a $k$-dimensional simplex $\sigma$ is defined as the convex hull of this set \cite{intro_to_TDA}.

A simplicial complex $\mathcal{K}$ (referred to as \textit{a geometric complex}) is a finite set of simplices $\{\sigma_i\}_{i=1}^n$ satisfying the following properties \cite{intro_to_TDA}:
\begin{itemize}
    \item Every face $\Delta \subset \sigma_i$ is an element of $\mathcal{K}$.
    \item The intersection of any two simplices $\sigma_i \cap \sigma_j$ is either empty or their common face $\Delta \in \mathcal{K}$.
\end{itemize}

Similarly, in abstract spaces without assigned geometry, we define an \textit{abstract} simplicial complex:

An abstract simplicial complex $\mathcal{K}$ is a pair $(V,F)$, where $V$ is a set of vertices, and $F$ is a family of subsets of $V$ such that for any element $\sigma \in F$, all of its subsets belong to $V$ \cite{intro_to_TDA}.

It can be observed that a geometric simplicial complex corresponds to an abstract complex, defined by a combinatorial description of its vertices. Conversely, any abstract simplicial complex can be represented as a geometric complex in some Euclidean space -- this is referred to as the \textit{geometric realization} \cite{geom_realization}.

\begin{figure}[h!]
    \centering
\includegraphics[width=0.7\textwidth]{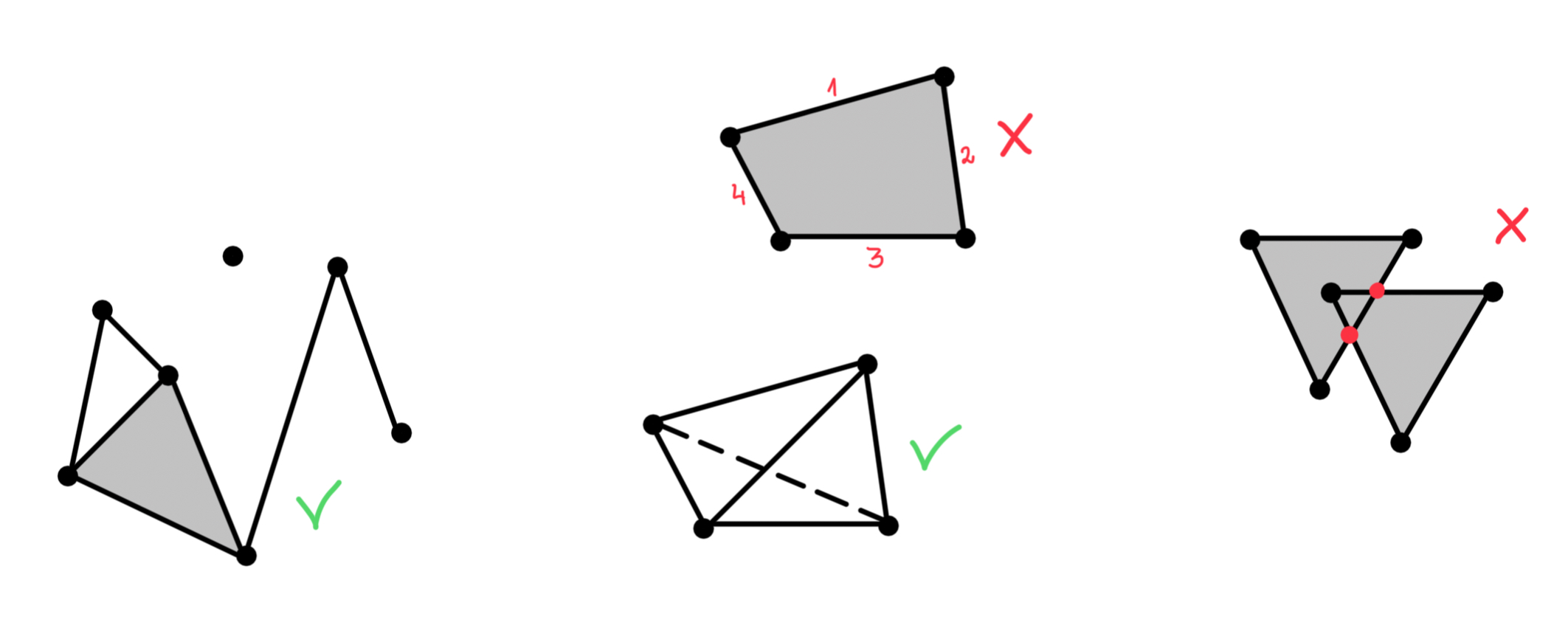}
    \caption{Examples of correctly and incorrectly constructed simplicial complexes.}
    \label{fig: przyklady kompl sympl}
\end{figure}

In the context of TDA, simplicial complexes provide us with a view of the connections between the analyzed data. More precisely, each point in the analyzed data is assigned a point in $\mathbb{R}^n$.
Given a point cloud, defined as a collection of data points within a metric space, we can construct a complex where the vertices are points representing the analyzed data, and simplices are created according to a certain rule that measures the similarity of the data.

\subsection{Vietoris-Rips and Čech Complexes}
The Vietoris-Rips complex $\mathcal{VR}(V)$ is a simplicial complex constructed on a set of points $V$ in a metric space $(M,d)$. For a fixed $\alpha > 0$, a simplex $\sigma \in \mathcal{VR}(V)$ is defined as a subset of $V$ such that $d(x_i,x_j) \le \alpha, \  \forall{x_i,x_j \in \sigma}$ \cite{intro_to_TDA}. In other words, for \( \epsilon \geq 0 \), the Vietoris-Rips complex is defined as:
\[ \mathcal{VR}(X, \epsilon) = \{ \sigma \subset X \mid \forall x, y \in \sigma, \; d(x, y) \leq \epsilon \} \] \label{VR}
where \( d \) is the metric in the space \( \mathbb{R}^n \).

In a similar manner, we define the Čech complex $\mathcal{C}(V)$. 
Given a set $V$ of points in a metric space $(M, d)$, a simplex $\sigma \in \mathcal{C}(V)$ is defined as a subset of $V$ such that the closed balls centered at $x \in \sigma$ have a non-empty intersection \cite{intro_to_TDA}: 
$$
\bigcap_{x\in \sigma} \overline{B_{\frac{\alpha}{2}}(x)} \not = \emptyset
$$

These constructions are frequently used when applying TDA methods in data analysis \cite{use_of_tda_1} \cite{use_of_tda_2}.

\begin{figure}[h!]
    \centering
    \includegraphics[width=0.7\textwidth]{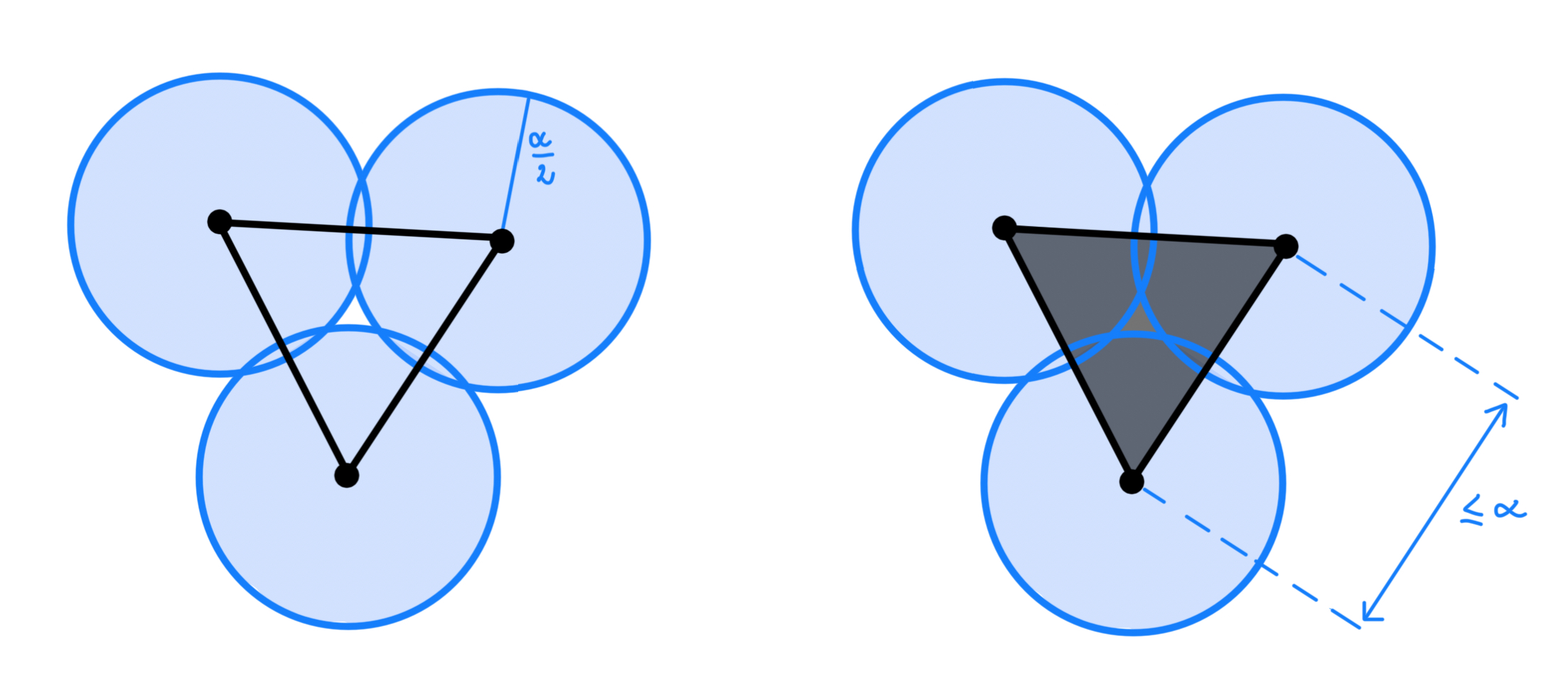}
    \caption[Construction of Vietoris-Rips and Čech complexes]{Construction of Vietoris-Rips and Čech complexes. The Čech complex (on the left) consists only of 1-simplices because the closed balls do not have a common intersection. The Vietoris-Rips complex (on the right) contains a 2-simplex because the distance between all points is no greater than $\alpha$.}
    \label{fig: VR i Cech kompleksy}
\end{figure}

\subsection{Cubical Complexes} \label{sec: cubical complexes}
Analogously to simplicial complexes, cubical complexes are defined, in which instead of simplices, objects generalizing points, line segments, squares, and cubes in any dimension are used \cite{KMM}.

We define an elementary interval as a closed interval $I \subset \mathbb{R}$:
\begin{equation} \label{eq: elementary interval}
I = [a,a+1], \quad a\in \mathbb{Z}
\end{equation}
\begin{center} 
or
\end{center} 
\begin{equation} \label{eq: deg interval}
I=[a,a] \stackrel{def.}{=} [a]
\end{equation}

In the case of \eqref{eq: elementary interval}, the construction corresponds to a 1-dimensional standard simplex.
The case of \eqref{eq: deg interval}, on the other hand, corresponds to a 0-dimensional simplex (thus a point).

An elementary cube $Q$ is defined as the Cartesian product of $n$ intervals:
$$
Q = \bigtimes_{k=1}^{n} I_k = I_1 \times I_2 \times \ldots \times I_n 
$$

The embedding dimension of a cube is determined by the number of intervals used to define it. The dimension of the cube $Q$ is the number of non-degenerate intervals.

Similarly to simplicial complexes, we can introduce the concept of a cubical complex. Its use in analyzing digital images is particularly justified from the perspective of computational complexity --- an $n$-cube $Q$ can be described using only one vertex \cite{KMM}.

\subsection{Filtration}

A filtration of a complex $\mathcal{K}$ is defined as a family of complexes $\{\mathcal{K}_t\}_{t\in T}, \  T\subseteq \mathbb{R}$ with the following properties \cite{computing_PH}:
\begin{itemize}
    \item $\forall_{t\in T} \mathcal{K}_t \subset \mathcal{K}$

    \item $t < t' \implies \mathcal{K}_t \subset \mathcal{K}_{t'}$

    \item $\cup_{t\in T} \mathcal{K}_t = \mathcal{K}$

\end{itemize}

\begin{figure}[h!]
\begin{center}
    
\includegraphics[width=1\textwidth]{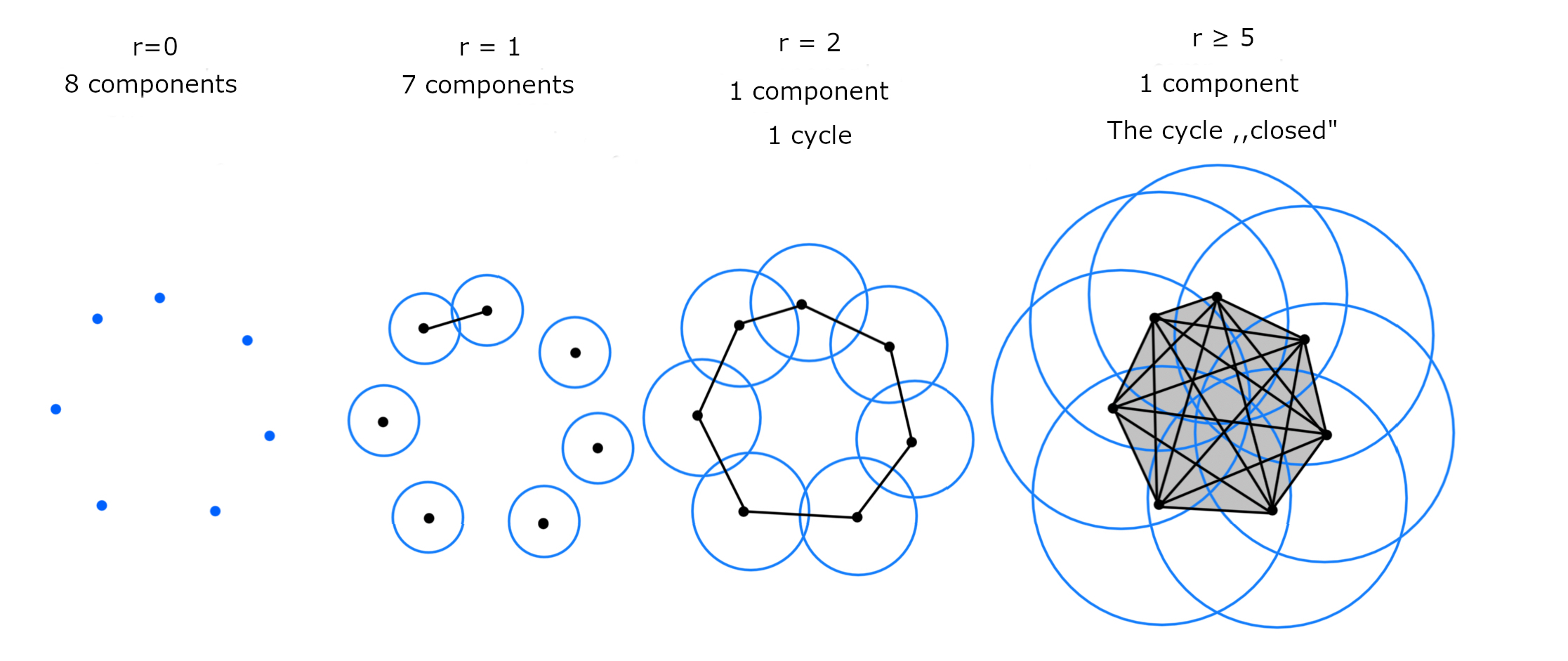}

    \caption[Filtration of the Vietoris-Rips complex for a range of radii $r$]{Filtration of the Vietoris-Rips complex for a range of radii $r$. We observe that for changing values of $r$, the topological properties of the analyzed space change — the number of connected components changes, and cycles appear and disappear.}
    \label{fig: filtracja vr}
\end{center}

\end{figure}

\subsection{Persistent homology and persistence diagrams}

Persistent homology is an extension of classical homology theory \cite{CLSHomology}, which analyzes the topological features of data across multiple scales. The main goal of persistent homology is to identify topological features that are stable and significant at different levels of detail \cite{betti}. Persistent homology involves analyzing topological spaces that change with a scalar parameter \( \epsilon \). A filtration of topological spaces is a sequence of subspaces:
\[ \emptyset = X_0 \subseteq X_1 \subseteq X_2 \subseteq \ldots \subseteq X_n = X \]

For each space \( X_i \), we compute the homology groups \( H_k(X_i) \) \cite{computing_PH}. If \( X_i \subseteq X_j \), there exists an induced homomorphism on the homology groups:
\[ f_{i,j}^k: H_k(X_i) \to H_k(X_j) \]

Topological features, such as connected components or cycles, may appear and disappear as \( \epsilon \) changes. The moment a feature appears is called its "birth" \ (\( \epsilon_b \)), and the moment it disappears is called its "death" \ (\( \epsilon_d \)). Persistent homology tracks these moments, forming pairs \((\epsilon_b, \epsilon_d)\). A persistence diagram is a visual representation of the pairs \((\epsilon_b, \epsilon_d)\). On the horizontal axis, the values of \(\epsilon_b\) (births) are plotted, and on the vertical axis, the values of \(\epsilon_d\) (deaths) are plotted. A point in the diagram closer to the diagonal represents short-lived features, while points farther from the diagonal correspond to longer-lasting, topologically significant features.

Persistence diagrams are a graphical tool used to present persistent homology. The process of creating persistence diagrams consists of a few steps: constructing a filtration of topological spaces, calculating the homology for each space in the filtration, and identifying the birth and death moments of topological features. A fundamental step in creating persistence diagrams is the construction of a filtration. 

Suppose we have a finite set of points \( X \subset \mathbb{R}^n \). In the simplest case, the filtration can be defined as a sequence of Vietoris-Rips complexes \ref{VR} constructed by the radius \( \epsilon \) chosen from an increasing sequence of values. In each space \( \mathcal{VR}(X, \epsilon) \), we can compute the homology groups \( H_k \), which represent $k$-dimensional topological features, such as connected components (\( H_0 \)), cycles and 1-dimensional "holes" \ (\( H_1 \)), and, in general, $k$-dimensional voids (\( H_k \)). We analyze the changes in the homology groups as \( \epsilon \) increases. A non-trivial homology generator (e.g., a cycle) may "be born" \ at a certain value \( \epsilon_b \) and "die" \ at a value \( \epsilon_d \). A persistence diagram is a collection of such pairs \( (\epsilon_b, \epsilon_d) \), where each pair represents a specific topological feature of the data.

\begin{figure}
    \centering
    \includegraphics[width=1\textwidth]{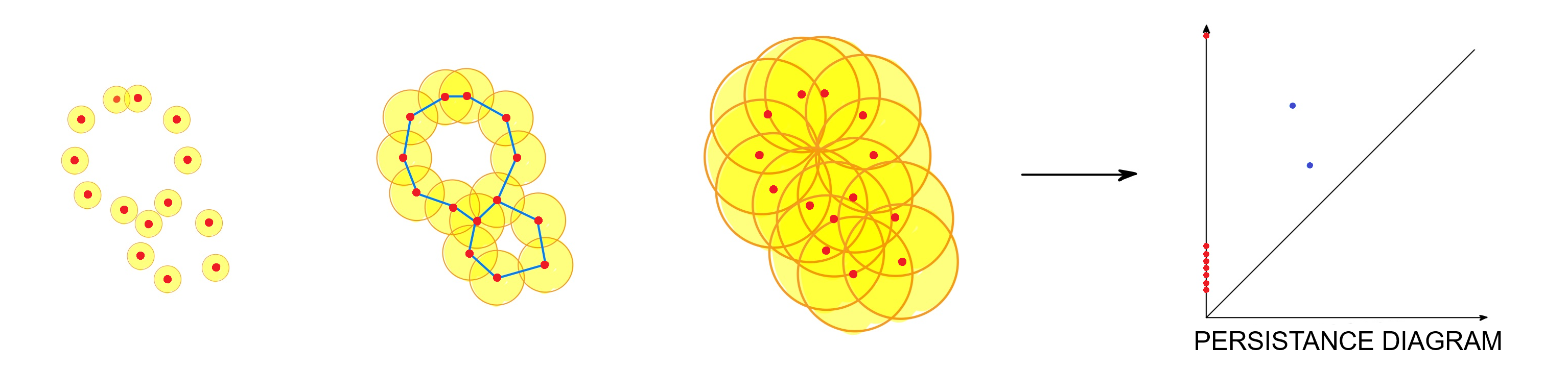}
    \caption{Construction of a persistence diagram based on the Vietoris-Rips filtration.}
    \label{fig: diagram na podst filtr}
\end{figure}

\subsection{Vector representation of persistence diagrams} \label{sec: definicja_PI}

Persistence diagrams carry valuable topological information about the analyzed space. There are methods for comparing diagrams, such as the \textit{bottleneck} and \textit{Wasserstein} metrics -- however, their computational cost increases significantly as the number of points in the diagram grows \cite{koszt_obliczeniowy_PD}. Due to this, and the need to represent persistence diagrams in a way that allows the use of topological information in machine learning processes, various vectorization methods have been introduced. These include persistence landscapes, Betti curves, persistence silhouettes, and persistence images \cite{PI} \cite{landscapes_introduced} \cite{Landscapes} \cite{koszt_obliczeniowy_PD} \cite{porownanie_wektoryzacji}.

In this work, we will use persistence images (PI) as the primary tool for representing topological information in machine learning processes.

\begin{figure}[h!]
    \centering
    \includegraphics[width=0.9\textwidth]{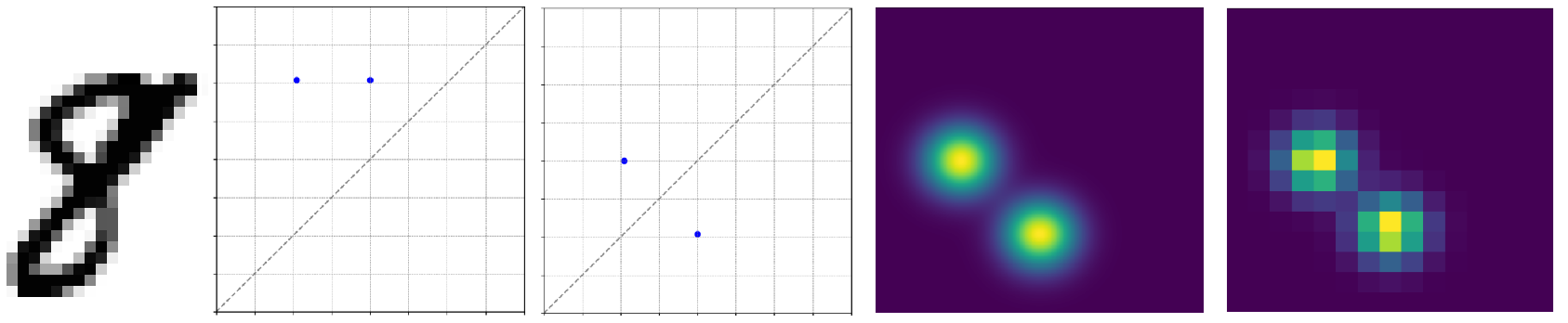}
    \caption{The process of creating persistence images.}
    \label{fig: proces tworzenia PI}
\end{figure}

Adams et al. \cite{PI} define a persistence image as:

$$
\rho_B(x,y) = \sum_{u \in T(B)} f(u) \Phi_u(x,y)
$$
\label{pers_img}
where:
\begin{itemize}
    \item $B$ represents a persistence diagram.
    \item $T$ denotes the transformation of the diagram $B \ni (x,y) \mapsto (x,y-x) \in T(B)$.
    \item $f$ is a weighting function.
    \item $\Phi_u$ is a two-dimensional normal distribution centered at point $u$.
\end{itemize}
To obtain an image in the form of a vector in a finite-dimensional space, we further divide it into pixels. Thus, we partition the region $\rho_B$ into $n\times n$ pixels, assigning each one the value:
$$
I(\rho_B)_p = \iint_p \rho_B(x,y) dxdy
$$
The process described above is illustrated in Figure \ref{fig: proces tworzenia PI}.

\section{Integration of TDA with convolutional neural networks} \label{sec: models}

To investigate the effectiveness of TDA methods, we will compare the results of three convolutional neural network models. These will be architectures with a similar number of parameters, differing in input information.

\begin{enumerate}
    \item A CNN operating on raw grayscale images (referred to in this work as the \textit{raw} model).
    \item A CNN utilizing preprocessed images in the form of persistence images (referred to in this work as the TDA model).
    \item A network operating on input data consisting of "stitched" images and persistence images. This model will be described in more detail in subsection \ref{sec: vector_stitching} (referred to in this work as the \textit{Vector-Stitching} model).
\end{enumerate}

In the process of training and validating the models, we use the standard MNIST dataset \cite{mnist} containing hand-drawn digits from 0 to 9. For the purposes of the experiment, we modify the dataset by adding two types of noise to each image: white noise and salt-and-pepper noise.

\subsection{TDA \textit{Pipeline}} \label{sec: pipelineTDA}
To preprocess the input data into persistence images, we use the \textit{giotto-tda} library \cite{giotto_github} \cite{tauzin2020giottotda}. The \textit{giotto-tda} library is focused on using TDA methods in machine learning processes, unlike other available libraries (e.g., GUDHI \cite{gudhi}), which offer more tools for topological data analysis but require deeper integration with neural networks.

\begin{figure}[h!]
    \centering
    \includegraphics[width=0.93\textwidth]{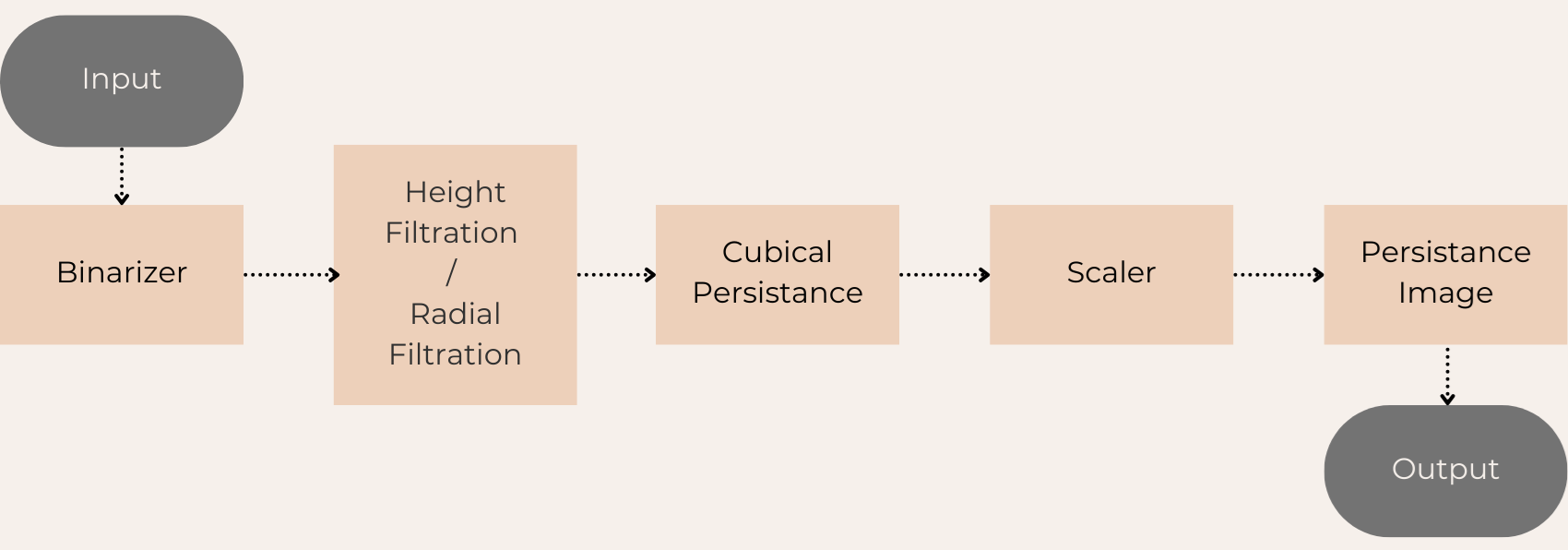}
    \caption[Input data processing pipeline diagram.]{Input data processing pipeline diagram.}
    \label{fig: schemat pipelinu}
\end{figure}
\newpage
\begin{itemize}
    \item \textbf{Binarization}: We convert the input images to a binarized form -- each pixel is assigned a value of 1 if its value in the original image exceeds a fixed threshold (we use a threshold of 0.4), otherwise, it is assigned the value of 0. We denote the binarized image as $B$. 
    
    \item \textbf{HeightFiltration}: We use 8 different filtration directions, during which each pixel $p \in B$ is assigned the value:
    $$
    H(p) = 
    \begin{cases}
        \langle p, v \rangle & \text{if } B(p) = 1 \\
        H_\infty & \text{if } B(p) = 0
    \end{cases}
    $$
    where $v$ represents the direction of the filtration -- in the case of two-dimensional images, it is a vector from the space $\mathbb{R}^2$. $\langle p,v \rangle$ denotes the distance of point $p$ from the hyperplane defined by $v$; $H_\infty$ is the filtration value of the pixel farthest from the considered hyperplane \cite{TDA_MNIST_READING}.
    \item \textbf{RadialFiltration}: We use 9 different filtration points (centers). Each pixel $p \in B$ is assigned its Euclidean distance from the filtration point.
    $$
    R(p) = 
    \begin{cases}
        d_e(c,p) & \text{if } B(p) = 1 \\
        R_\infty & \text{if } B(p) = 0
    \end{cases}
    $$
    where $c$ is the filtration point (center). Similar to \textit{HeightFiltration}, here we also set $R_\infty$ as the filtration value of the pixel furthest from point $c$ \cite{TDA_MNIST_READING}.

    \item \textbf{CubicalPersistence}: We use a module that generates the persistence diagram using cubical complex filtration.

    \item \textbf{Scaler}: To obtain uniform topological information about the analyzed images, we scale the persistence diagrams.

    \item \textbf{PersistenceImage}: After processing the persistence diagram according to the steps described above, we generate the PI as described in section \ref{sec: definicja_PI}.
\end{itemize}

\begin{figure}[h!]
    \centering
    \includegraphics[width=0.9\textwidth]{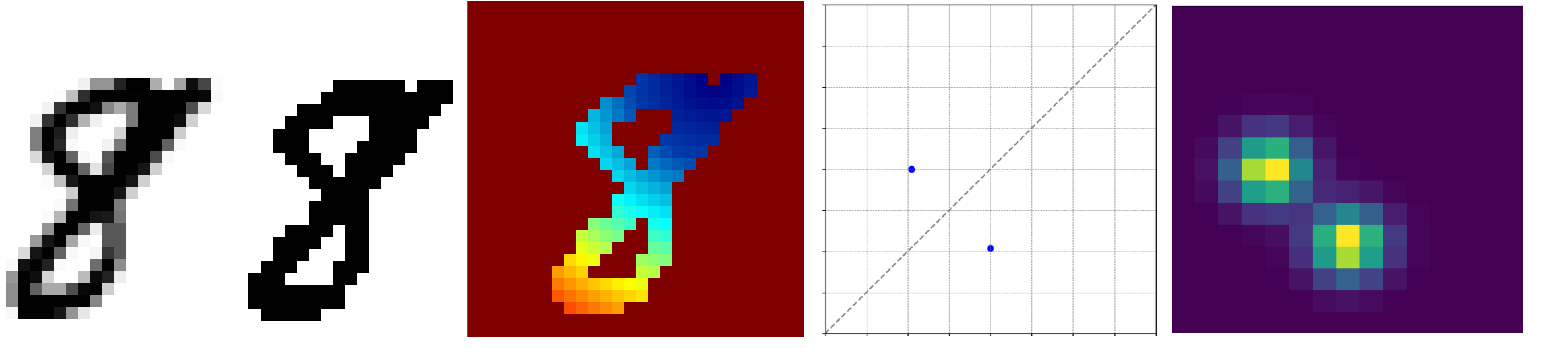}
    \caption{Image processing through the TDA \textit{pipeline}.}
    \label{fig: pipiline tda na przykladzie}
\end{figure}
\newpage
The described pipeline generates a persistence image consisting of 34 layers, where each layer corresponds to a separate persistence image calculated for one of 17 filtrations, successively for $H_0$ and $H_1$.

\subsection{\textit{Vector Stitching} method}\label{sec: vector_stitching}

\begin{figure}[h!]
    \centering
    \includegraphics[width=0.5\textwidth]{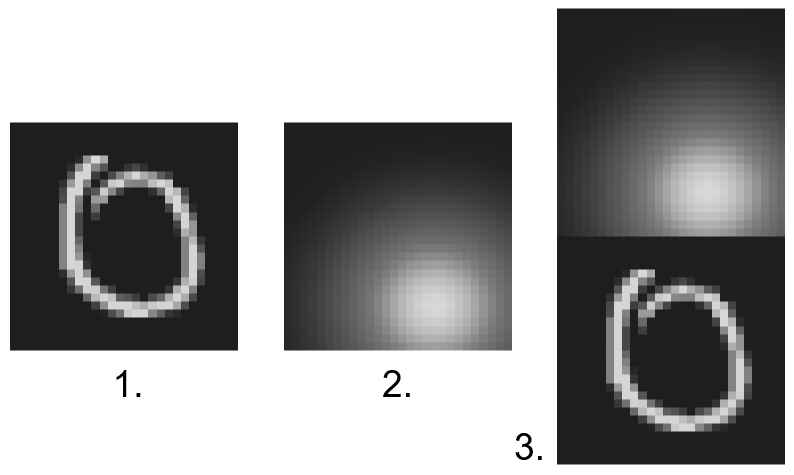}
    \caption{The method of combining the original image with its corresponding persistence image. 1. Raw image 2. Persistence diagram 3. Stitched images.}
    \label{fig: obraz VS}
\end{figure}

The third model presented in section \ref{sec: models} is a convolutional neural network operating on "stitched" \ images. McGuire et al. \cite{Do_neural_} show that neural networks trained on raw data have a limited ability to capture the topological properties of the analyzed images. To fully utilize the information contained in the analyzed data, we combine the raw image with the corresponding persistence image described in \mbox{section \ref{sec: pipelineTDA}}. The result of such stitching is illustrated in the example shown in Figure \ref{fig: obraz VS}.

\section{Experiments} \label{sec: eksperyment}

The aim of the conducted experiment is to evaluate the impact of incorporating TDA methods on the accuracy of CNNs in image recognition tasks under conditions of limited training data. Additionally, the experiment investigates how the integration of topological information influences the robustness of CNNs to noise, aiming to provide insights into the potential advantages of architectures enhanced with TDA in deep learning applications.

\subsection{Datasets} \label{sec: zbiory_danych}
The experiment was conducted on the standard MNIST dataset \cite{mnist}, which contains hand-written digits from 0 to 9. For the purposes of the experiment, the dataset was modified by adding two types of noise: white noise and salt-and-pepper noise. We deliberately utilized small datasets, with images selected in a pseudorandom manner, to illustrate the extent of accuracy improvement achieved through the \textit{Vector Stitching} method when training with a limited number of samples. To ensure the reliability of our results and to draw statistically significant conclusions, these training and test sets were chosen 10 times without duplication for each experiment run. From this modified data, we created training and test sets for two experiments:
\begin{enumerate}
    \item Training models on clean data and validating on noisy data only. \\
    The training sets have the following sizes: 100, 250, 500, and 1000 images, and the test sets contain 100 images each.
    
    \item Training models on a mix of clean and distorted data in a 90/10 ratio, with validation on noisy data only. \\
    The training sets have the following sizes: 100, 250, 500, and 1000 images, and the test sets contain 100 images each.
\end{enumerate}

\subsection{Models} \label{sec: modele}

For our experiments, we utilized the TensorFlow library to construct three distinct CNN architectures. The first model (\textit{RAW model}) was trained on raw data, serving as a baseline for performance comparison. The second model (\textit{TDA model}) was trained exclusively on vectorized topological features derived through Persistence Images \ref{pers_img}, allowing us to assess how well a model can perform when trained solely on these features. The third model (\textit{Vector Stitching model}) was trained on stitched data, which combines raw data with TDA-filtered samples, aiming to leverage the complementary strengths of both inputs. To ensure a fair comparison, we designed all three models to have as similar a total number of parameters as possible. Minor variations in parameter counts arose due to differences in input sizes for stitched data samples.

\section{Results of the experiments} \label{sec: wyniki i analiza}

In this section, we show the results of our experiments conducted with the two datasets described in Section \ref{sec: zbiory_danych} and the three models discussed in Section \ref{sec: modele}.

\subsection{Results of the network trained on clean training data}

\begin{figure}[h!]
    \centering
    \includegraphics[width=1\textwidth]{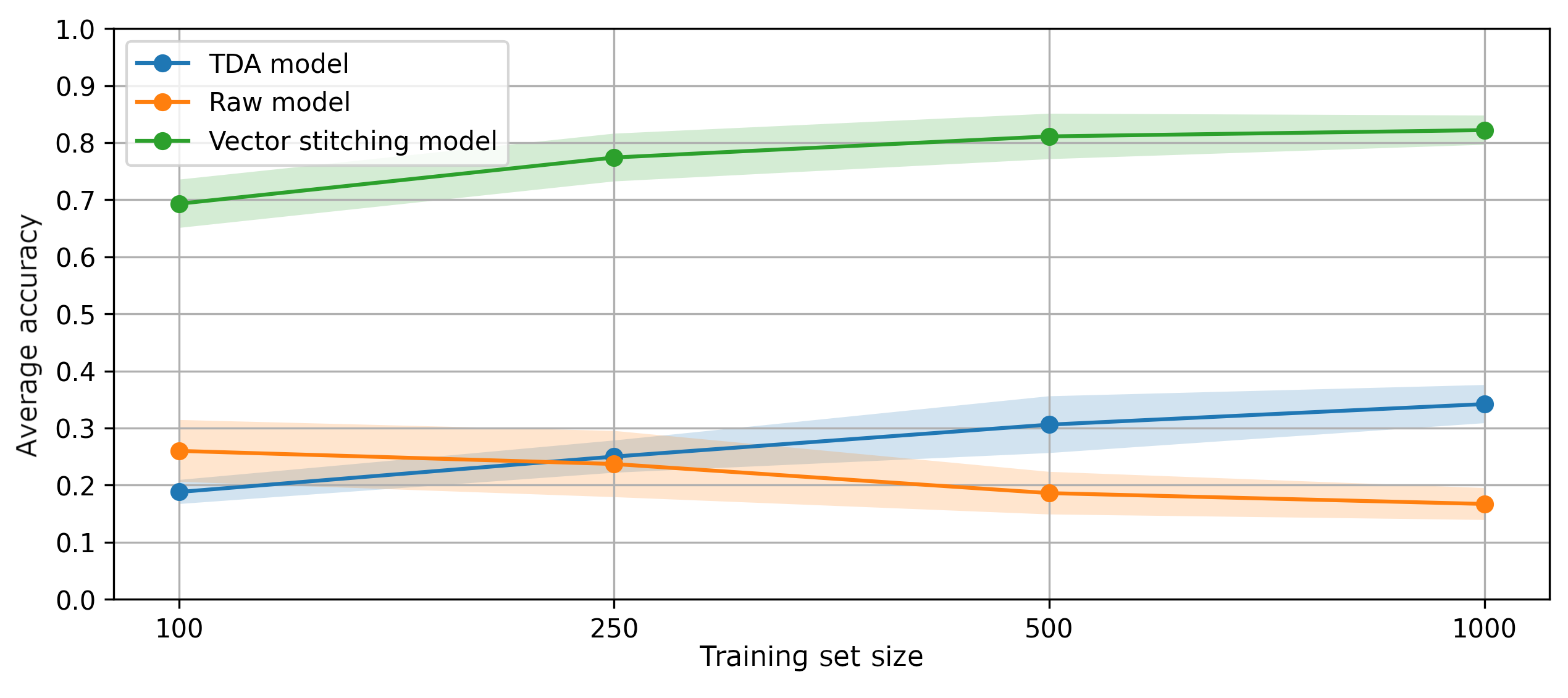}
    \caption{Results of the three models trained on clean training sets with confidence intervals.}
    \label{fig: wykresy_czyste}
\end{figure}

Table \ref{tab: podsumowanie_tda_czyste} shows the performance results of models trained on training sets consisting solely of clean data, followed by validation on noisy data sets. The average accuracy, standard deviation, and 95\% confidence intervals calculated from 10 iterations of the experiment are shown. The results are visualized in Figure \ref{fig: wykresy_czyste}.

\begin{table}[h!]
\centering
\begin{tabular}{@{}ccccc@{}}
\toprule
Model & Training Set Size & Accuracy & Confidence Interval \\
\midrule
TDA & 100 & 0.1880 $\pm$ 0.0340 & (0.1669, 0.2091) \\
TDA & 250 & 0.2500 $\pm$ 0.0456 & (0.2217, 0.2783) \\
TDA & 500 & 0.3060 $\pm$ 0.0803 & (0.2562, 0.3558) \\
TDA & 1000 & 0.3420 $\pm$ 0.0540 & (0.3085, 0.3755) \\
\midrule
RAW & 100 & 0.2600 $\pm$ 0.0874 & (0.2058, 0.3142) \\
RAW & 250 & 0.2370 $\pm$ 0.0936 & (0.1790, 0.2950) \\
RAW & 500 & 0.1860 $\pm$ 0.0602 & (0.1487, 0.2233) \\
RAW & 1000 & 0.1670 $\pm$ 0.0452 & (0.1390, 0.1950) \\
\midrule
V-S & 100 & 0.6930 $\pm$ 0.0683 & (0.6507, 0.7353) \\
V-S & 250 & 0.7740 $\pm$ 0.0676 & (0.7321, 0.8159) \\
V-S & 500 & 0.8110 $\pm$ 0.0643 & (0.7712, 0.8508) \\
V-S & 1000 & 0.8220 $\pm$ 0.0414 & (0.7963, 0.8477) \\
\bottomrule
\end{tabular}
\vspace{0.2cm}
\caption{Summary of results for the TDA, RAW and Vector Stitching models trained on successive training sets consisting solely of clean data.}
\label{tab: podsumowanie_tda_czyste}
\end{table}

\subsection{Results of models trained on clean and noisy training data}

\begin{figure}[h!]
    \centering
    \includegraphics[width=1\textwidth]{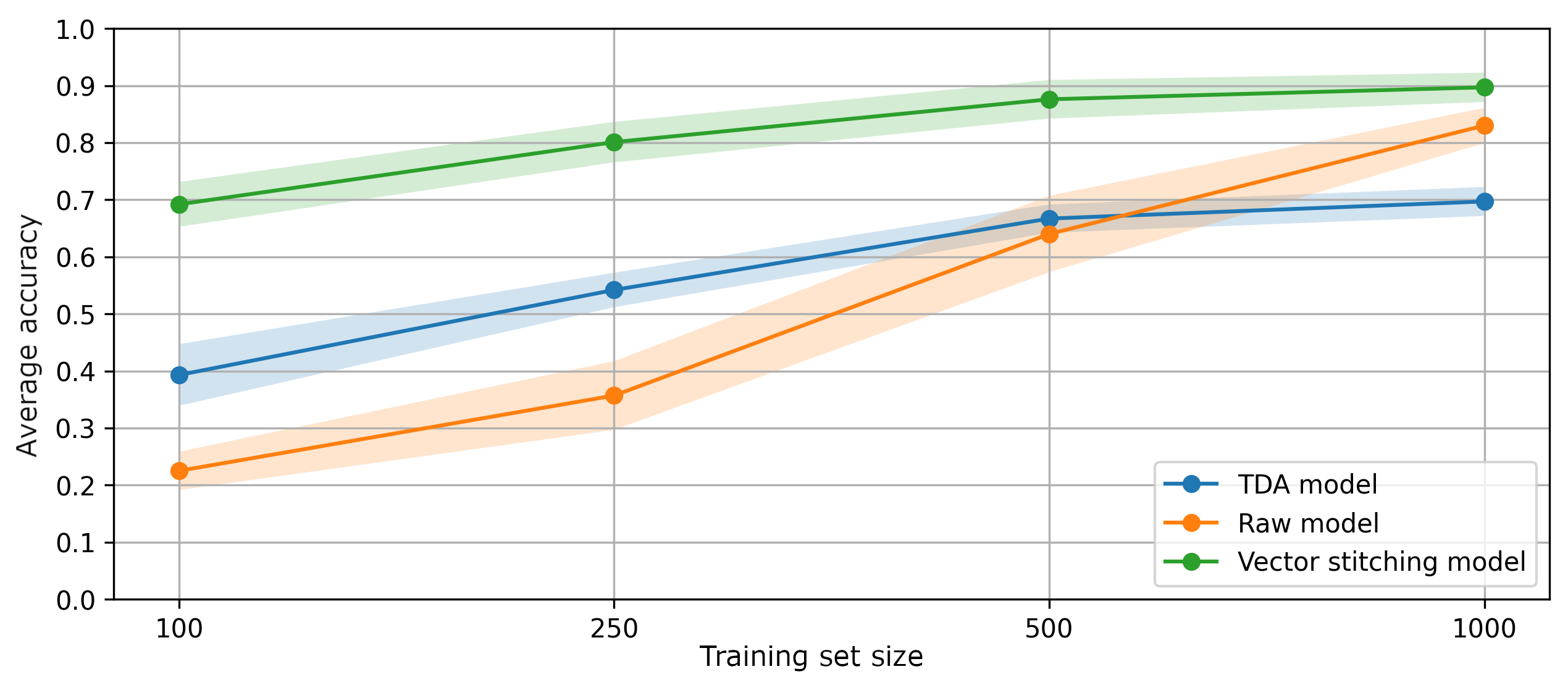}
    \caption{Results of the three models trained on training sets with a 90/10 split between clean and noisy data, with confidence intervals shown.}
    \label{fig: wykresy_zaszumione}
\end{figure}

Tables \ref{tab: podsumowanie_zaszumione_tda} shows the performance results of models trained on training sets consisting of 90\% clean images and 10\% noisy images. All models were then tested on sets consisting only of noisy images. The average accuracy, standard deviation, and 95\% confidence intervals calculated from 10 iterations of the experiment are shown. The results are visualized in Figure \ref{fig: wykresy_zaszumione}.

\begin{table}[h!]
\centering
\begin{tabular}{@{}ccccc@{}}
\toprule
Model & Training Set Size & Accuracy & Confidence Interval \\
\midrule
TDA & 100 & 0.3930 $\pm$ 0.0871 & (0.3390, 0.4470) \\
TDA & 250 & 0.5420 $\pm$ 0.0487 & (0.5118, 0.5722) \\
TDA & 500 & 0.6670 $\pm$ 0.0400 & (0.6422, 0.6918) \\
TDA & 1000 & 0.6970 $\pm$ 0.0410 & (0.6716, 0.7224) \\
\midrule
RAW & 100 & 0.2250 $\pm$ 0.0541 & (0.1915, 0.2585) \\
RAW & 250 & 0.3570 $\pm$ 0.0971 & (0.2968, 0.4172) \\
RAW & 500 & 0.6400 $\pm$ 0.1081 & (0.5730, 0.7070) \\
RAW & 1000 & 0.8300 $\pm$ 0.0498 & (0.7991, 0.8609) \\
\midrule
V-S & 100 & 0.6920 $\pm$ 0.0634 & (0.6527, 0.7313) \\
V-S & 250 & 0.8010 $\pm$ 0.0572 & (0.7656, 0.8364) \\
V-S & 500 & 0.8760 $\pm$ 0.0546 & (0.8421, 0.9099) \\
V-S & 1000 & 0.8970 $\pm$ 0.0415 & (0.8713, 0.9227) \\
\bottomrule
\end{tabular}
\vspace{0.2cm}
\caption{Summary of results for the TDA, RAW and Vector Stitching models trained on successive training sets consisting of clean/noisy data in a 90/10 ratio.}
\label{tab: podsumowanie_zaszumione_tda}
\end{table}

\newpage

\begin{figure}[h!]
    \centering
    \includegraphics[width=0.90\textwidth]{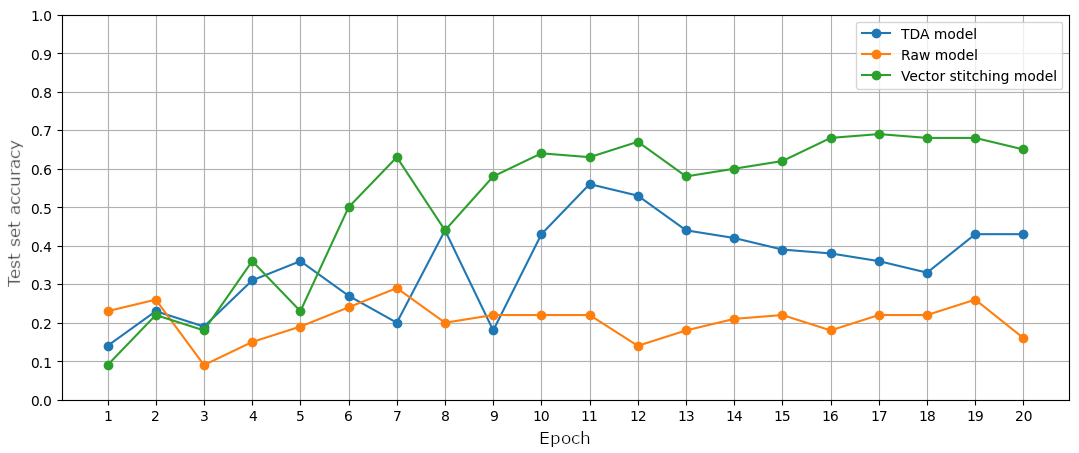}
    \caption{Accuracy results for one of the iterations, epoch by epoch. The training set consists of a mixture of 100 clean and noisy images.}
    \label{fig: ebe100}
\end{figure}

\begin{figure}[h!]
    \centering
    \includegraphics[width=0.90\textwidth]{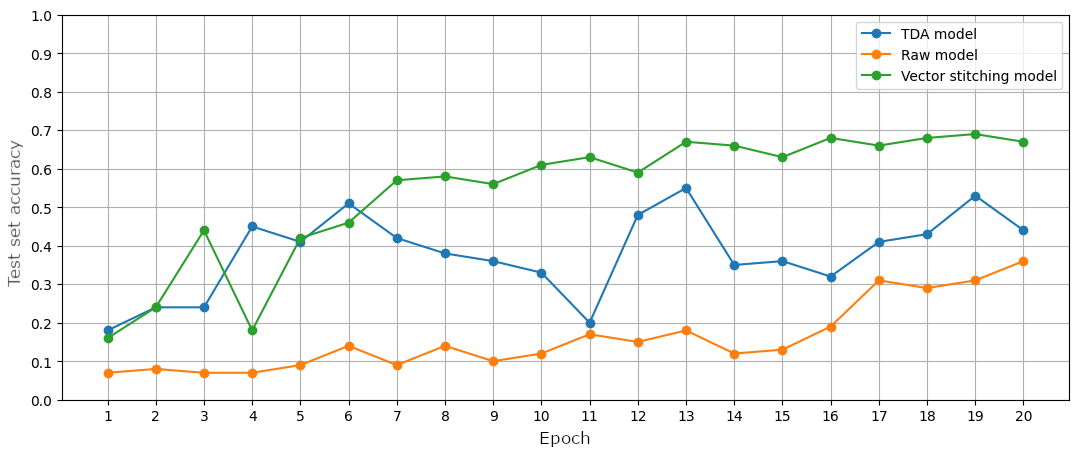}
    \caption{Accuracy results for one of the iterations, epoch by epoch. The training set consists of a mixture of 250 clean and noisy images.}
    \label{fig: ebe250}
\end{figure}

\begin{figure}[h!]
    \centering
    \includegraphics[width=0.9\textwidth]{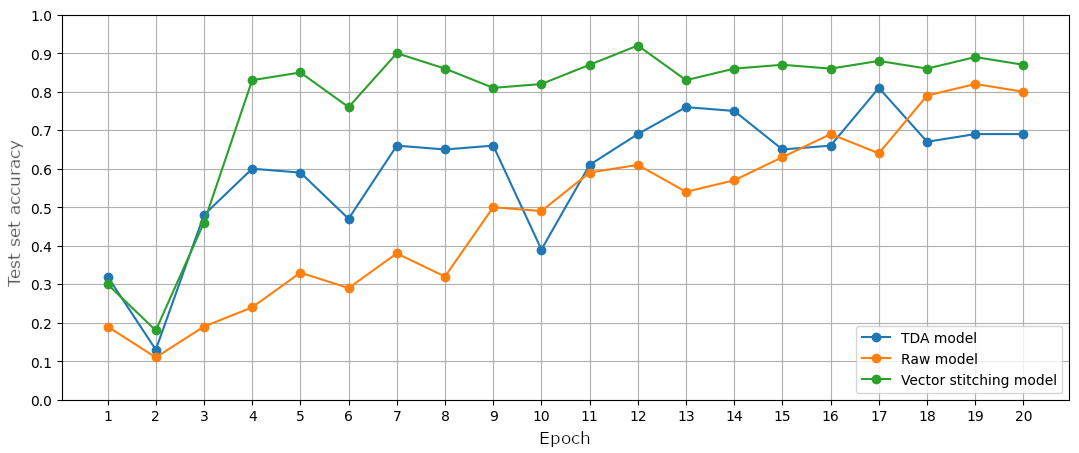}
    \caption{Accuracy results for one of the iterations, epoch by epoch. The training set consists of a mixture of 500 clean and noisy images.}
    \label{fig: ebe500}
\end{figure}

\begin{figure}[h!]
    \centering
    \includegraphics[width=0.90\textwidth]{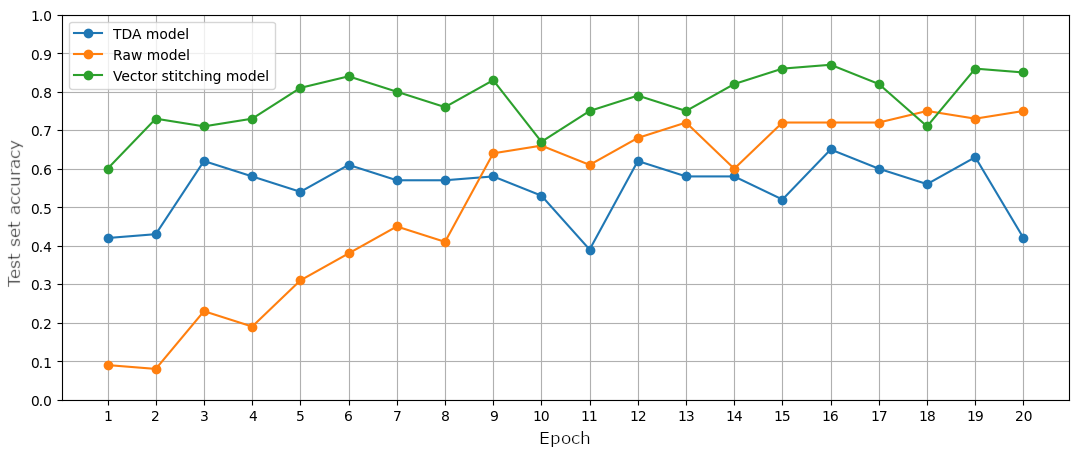}
    \caption{Accuracy results for one of the iterations, epoch by epoch. The training set consists of a mixture of 1000 clean and noisy images.}
    \label{fig: ebe1000}
\end{figure}

\newpage

\subsection{Discussion of the results}

The results of the experiments presented in Table \ref{tab: podsumowanie_tda_czyste} show that the \textit{Vector-Stitching} method, which combines raw images with persistence images, achieves better results compared to the standard model and the model using only PI. In the case of purely clean training data, the \textit{Vector-Stitching} model achieved 70\% accuracy with the smallest training set (100 images), and the best result (82\%) with the largest training set (1000 images). For this type of training and test data, the \textit{Raw} model was unable to generalize information extracted from clean images to classify noisy images. The results of the PI-based model showed an upward trend as the size of the training set increased, but it only reached around 30\% accuracy.

A similar trend can be observed with mixed training data, where the \textit{Vector-Stitching} model again outperformed the others, achieving 90\% accuracy with the largest training set. The other models for this training set reached the following accuracies: 70\% (PI) and 83\% (\textit{Raw}). With this type of training data, the \textit{Raw} model performed much better, surpassing the TDA model with 500 training images, but still fell a few percentage points short of the \textit{vector-stitching} model.

Analyzing the epoch-by-epoch results of the models, presented in Figures \ref{fig: ebe100} -- \ref{fig: ebe1000}, we can see that methods based on topological information reach high accuracy much faster than the standard method. For the training set with 1000 images (Fig. \ref{fig: ebe1000}), in epochs 1–9, the \textit{Vector-Stitching} model showed the highest accuracy, reaching over 60\% accuracy as early as in the first epoch. In the same epochs, the TDA model achieved accuracies on average 20 percentage points lower, while the \textit{Raw} model reached accuracies below 50\%, although it showed an upward trend as the epochs progressed. It was not until the 9th epoch that the \textit{Raw} model overtook the TDA model, but it still remained less accurate than the \textit{Vector-Stitching} model. The 1000-image training set was the only one where the \textit{Raw} model outperformed the TDA model. In all other, smaller training sets, the topological models consistently showed greater effectiveness than the standard model.

\section{Discussion} \label{sec: summary}

The above results suggest that the additional topological information provided by persistence images can significantly improve the performance of CNN models in image classification tasks, especially in the presence of noise in the data. The conducted experiments demonstrated that the integration of topological data analysis with convolutional neural networks contributes to the improvement of image classification accuracy. The "stitching" \ method proved to be the most effective both on clean and noisy data, confirming the potential of the proposed approach for more effective information extraction from data. It is important to note the significant differences in the performance of CNN models depending on the method used for processing the input data. In particular, the \textit{Vector Stitching} method, which combines raw images with persistence images, achieved better results compared to the \textit{raw} and TDA models, both on clean and noisy training data. The results of the experiments on clean data indicate that the \textit{Vector Stitching} method clearly outperforms other methods for all training data sets. This difference suggests that the additional information about topological properties provided by persistence images can significantly improve the performance of CNN models in image classification tasks, especially in the presence of data noise. A similar trend can be observed with mixed data, where the mentioned method also achieves better results than the others. This indicates the effectiveness of combining raw data with persistence images in improving model generalization, even with a limited amount of training data.

\subsection{Information-preserving transformations}

In the case of our findings, the \textit{vector stitching} method, which consists of a \textit{pipeline} combining raw data with filtered data using persistence images and the neural network itself, can be understood as a transformation designed to enhance information retention. In contrast, directly passing raw data to the neural network represents a more straightforward approach that does not incorporate topological features. The aim here is to highlight the distinction between transformations that preserve additional information and those that do not. While it may seem trivial to note that neural networks often do not fully utilize the information contained in the data, this work emphasizes the existence of certain transformations that reduce information loss between the data and the model's inference stage. Moreover, we identify such a transformation explicitly in section \ref{sec: pipelineTDA}. It is worth noting that simply increasing the number of neurons in the deep layers of the network could, in some cases, serve as an alternative approach to improving information retention.

\subsection{Significance in the context of existing research}

The results of our experiments are consistent with previous research indicating the benefits of integrating topological data analysis with convolutional neural networks. The studies by Adams et al. (2017) \cite{PI} and Bubenik and Dłotko (2017) \cite{landscapes_introduced} suggest that topological representations, such as persistence images, can provide stable and significant features that are difficult to detect with traditional image analysis methods. These solutions can be used, for instance, in the analysis of magnetic resonance or computed tomography images \cite{tda_in_medical}. There are also studies describing the potential use of TDA in time series analysis, such as detecting and analyzing gravitational waves, where the amount of noise and the required amount of training data make traditional methods inaccurate \cite{gravitational_waves}.

\subsection{Limitations and future research directions}

Although the \textit{Vector Stitching} method demonstrated high effectiveness, some limitations should be noted. First of all, the process of generating persistence images and integrating them with raw images requires significant computational resources, which can be a challenge in the context of real-time systems. Another major limitation is the Data Processing Inequality \cite{TI}. Regardless of the filtration or architecture we use, we will not be able to infer more information from the model than is already present in the data. In the case of the MNIST dataset, the amount of information contained in the samples is significantly limited, and thus the effectiveness the model can provide will also be limited. Therefore, it is likely that as the complexity of the problem and the amount of information in the data increases, methods like \textit{Vector Stitching} will be able to enhance the effectiveness of neural networks to a greater extent than classical methods. Another issue worth considering is the possibility that in many tasks where \textit{Vector Stitching} shows significant effects for a small number of network parameters, the effect observed in this work may diminish as the number of parameters increases.

Future research may focus on applying the proposed method to other types of visual data, such as medical images \cite{tda_in_medical}, where extremely precise classification and feature detection are required. Additionally, integrating TDA with other neural network architectures, such as transformers \cite{attention}, may open new perspectives for improving the performance of these models. There is also the matter of defining the properties of the transformations designed to enhance information retention. Developing a universal description for this class of transformations could significantly improve the explainability of neural networks.

\section*{Acknowledgments}

We would like to express our sincere gratitude to Associate Professor Paweł Pilarczyk for his support at every stage of the research, his attention to detail, and the long and insightful discussions.

%\bibliographystyle{plain}
%\bibliography{references}
%\printbibliography[heading=bibintoc,title={Bibliography}]

%\newpage
%\tableofcontents
%\listoffigures
%\listoftables

\end{document}